\documentclass{article}

\PassOptionsToPackage{numbers, compress}{natbib}

\usepackage[preprint]{neurips_2024}

\usepackage[utf8]{inputenc} 
\usepackage[T1]{fontenc}    
\usepackage{hyperref}       
\usepackage{url}            
\usepackage{booktabs}       
\usepackage{amsfonts}       
\usepackage{nicefrac}       
\usepackage{microtype}      
\usepackage{xcolor}         
\usepackage{amsmath}
\usepackage{graphicx}
\usepackage{makecell}
\usepackage{enumitem}
\usepackage{subcaption}
\usepackage{multirow}

\title{Efficient and Unbiased Sampling of Boltzmann Distributions via Consistency Models}

\author{
  Fengzhe Zhang\thanks{Equal Contribution}\\
  University of Cambridge\\
  \texttt{fz287@cam.ac.uk} \\
  \And
 Jiajun He$^*$\\
  University of Cambridge\\
  \texttt{jh2383@cam.ac.uk} \\
  \And
  Laurence I. Midgley\\
  University of Cambridge\\
  \texttt{lim24@cam.ac.uk} \\
   \And
Javier Antorán\\
  University of Cambridge\\
  \texttt{ja666@cam.ac.uk} \\
  \And
 José Miguel Hernández-Lobato\\
  University of Cambridge\\
  \texttt{jmh233@cam.ac.uk} \\
}

\begin{document}

\maketitle

\begin{abstract}
Diffusion models have shown promising potential for advancing Boltzmann Generators. However, two critical challenges persist: (1) inherent errors in samples due to model imperfections, and (2) the requirement of hundreds of functional evaluations (NFEs) to achieve high-quality samples. While existing solutions like importance sampling and distillation address these issues separately, they are often incompatible, as most distillation models lack the necessary density information for importance sampling.
This paper introduces a novel sampling method that effectively combines Consistency Models (CMs) with importance sampling. We evaluate our approach on both synthetic energy functions and equivariant $n$-body particle systems. Our method produces unbiased samples using only 6-25 NFEs while achieving a comparable Effective Sample Size (ESS) to Denoising Diffusion Probabilistic Models (DDPMs) that require approximately 100 NFEs.
\end{abstract}

\section{Introduction}
Sampling from Boltzmann distributions is a crucial task in statistical physics. 
Efficient sampling would enable the prediction of properties for new materials and drugs via computational simulations, reducing the need for costly experiments.
However, the high dimensionality and multimodal nature of these distributions pose significant challenges. Traditional methods like Monte Carlo Markov Chain (MCMC) \citep{metropolis1953equation, hastings1970monte} and Molecular Dynamics (MD) are often too time-consuming in complex settings.

Model-based Boltzmann generators \cite{noe2019boltzmann} offer an alternative approach to amortize the sampling process. Diffusion Models (DMs) \citep{sohl2015deep, ho2020denoising} have shown promise in this regard, generating high-quality samples by gradually denoising from random noise. However, DMs face two major drawbacks: slow sample generation and biased estimations due to discrepancies between the model and true distribution.

While combining DMs (specifically, DDPM \citep{ho2020denoising}) with Importance Sampling (IS) can address bias, it still requires hundreds of steps for a high Effective Sample Size (ESS).
On the other hand, distillation techniques \citep{salimansprogressive, song2023consistency, kim2023consistency} can significantly reduce sampling time. Yet, applying importance sampling to distilled models remains challenging due to the absence of an explicit model density.

In this work, leveraging recently developed Consistency Models (CMs) \citep{song2023consistency, kim2023consistency, li2024bidirectional}, we introduce an algorithm that significantly accelerates sampling while still supporting IS to correct the bias. Specifically, our contributions include:
\begin{itemize}[leftmargin=*]
\item A novel method integrating IS with Bidirectional Consistency Models (BCMs) to accelerate sampling. 
This approach alternates between deterministic steps along an ODE trajectory and stochastic steps via an SDE for both proposal and target distributions.
We use BCMs to accelerate the ODE part and the SDE to provide valid density for IS.
\item Introduction of E(3)-Equivariant CMs for molecular applications and extension of Consistency Trajectory Models (CTMs) to Bidirectional CTMs (BCTMs), which shows more accurate bidirectional traversal than BCMs on molecular applications.
\item Empirical verification on synthetic datasets and equivariant $n$-body systems, demonstrating unbiased sample production with only 6-25 NFEs, significantly outperforming the DDPM baseline.
\end{itemize}

\vspace{-4pt}
\section{Background}
\vspace{-3pt}
Before presenting our proposed method, we outline key preliminaries: Importance Sampling (IS), score-based Diffusion Models (DMs), and Bidirectional Consistency Models (BCMs).

\paragraph{Importance Sampling.}
To estimate integrals of the form $\mathbb{E}_{\boldsymbol{x} \sim p}[\phi(\boldsymbol{x})]$, where $p$ is a target distribution and $\phi$ is an evaluable function, Self-Normalized Importance Sampling (SNIS) is commonly used when direct sampling from $p$ is infeasible and only its unnormalized version $\bar{p}$ can be evaluated. Given a proposal distribution $q$ from which sampling is possible, the integral can be approximated as:
\begin{equation}
\mathbb{E}_{\boldsymbol{x} \sim p}[\phi(\boldsymbol{x})] \approx \frac{\sum_{n=1}^N w_n \phi(\boldsymbol{x}^{(n)})}{\sum_{n=1}^N w_n} = \sum_{n=1}^N \bar{w}_n \phi(\boldsymbol{x}^{(n)}),
\end{equation}
where $\boldsymbol{x}^{(n)} \sim q$, $w_n = \bar{p}(\boldsymbol{x}^{(n)}) / q(\boldsymbol{x}^{(n)})$ are importance weights, and $\bar{w}_n = w_n / \sum_{m=1}^N w_m$ are normalized weights. The effectiveness of the IS estimator is measured by the Effective Sample Size (ESS): $\widehat{\text{ESS}} = 1 / \sum_{n=1}^N \bar{w}_n^2$, with $1 \leq \widehat{\text{ESS}} \leq N$. SNIS provides asymptotically unbiased and consistent estimates, with bias and variance diminishing as the number of samples $N$ increases.

\paragraph{Score-based Diffusion Models.}
Diffusion Models (DMs) generate samples by gradually removing noise from Gaussian samples. This process can be formulated as solving a reverse Stochastic Differential Equation (SDE), such as in the Denoising Diffusion Probabilistic Model (DDPM) \cite{ho2020denoising}, or a Probability Flow (PF) Ordinary Differential Equation (ODE) \citep{song2020score}: $d\boldsymbol{x} = \left[\boldsymbol{f}(\boldsymbol{x}, t) - \frac{1}{2}g(t)^2 \nabla_{\boldsymbol{x}} \log p_t(\boldsymbol{x}) \right] dt$, where $t \in [0, T]$, $\boldsymbol{f}$ and $g$ are drift and diffusion coefficients of $\boldsymbol{x}_t$, and $p_t$ is the marginal density at time $t$.
Score-based DMs learn to approximate $\nabla \log p_t(\boldsymbol{x})$ using score matching \citep{hyvarinen2005estimation, song2019generative}. For sampling, this learned score function is used to solve the PF ODE from $T$ to $\epsilon$. From now on, we adopt the choices made by \cite{karras2022elucidating}, setting $T=80$, $\epsilon=0.002$, $\boldsymbol{f}(\boldsymbol{x}, t)=\boldsymbol{0}$ and $g(t)=\sqrt{2t}$.

\paragraph{Consistency Models.}
A significant limitation of DMs is their slow sampling speed, often requiring hundreds of numbers of functional evaluations (NFEs). Consistency Models (CMs) \citep{song2023consistency} address this issue by directly learning the integral of the PF ODE, which enables mapping any point $\boldsymbol{x}_t$ at time $t$ to the starting time $\epsilon$ along the same solution trajectory, allowing one-step or few-step sampling. Consistency Trajectory Models (CTMs) \citep{kim2023consistency} extend CMs by learning to traverse from $\boldsymbol{x}_t$ at any time $t$ to time $u$ along the denoising direction (i.e., $u \leq t$) on the same solution trajectory. This extension provides greater flexibility in balancing sample quality and NFEs. Bidirectional Consistency Models (BCMs) \citep{li2024bidirectional} further generalize the approach by enabling both forward and backward traversal along the trajectory, offering more versatility in both sampling and the inversion process.
\begin{figure}[t]
    \centering
    \includegraphics[width=\textwidth]{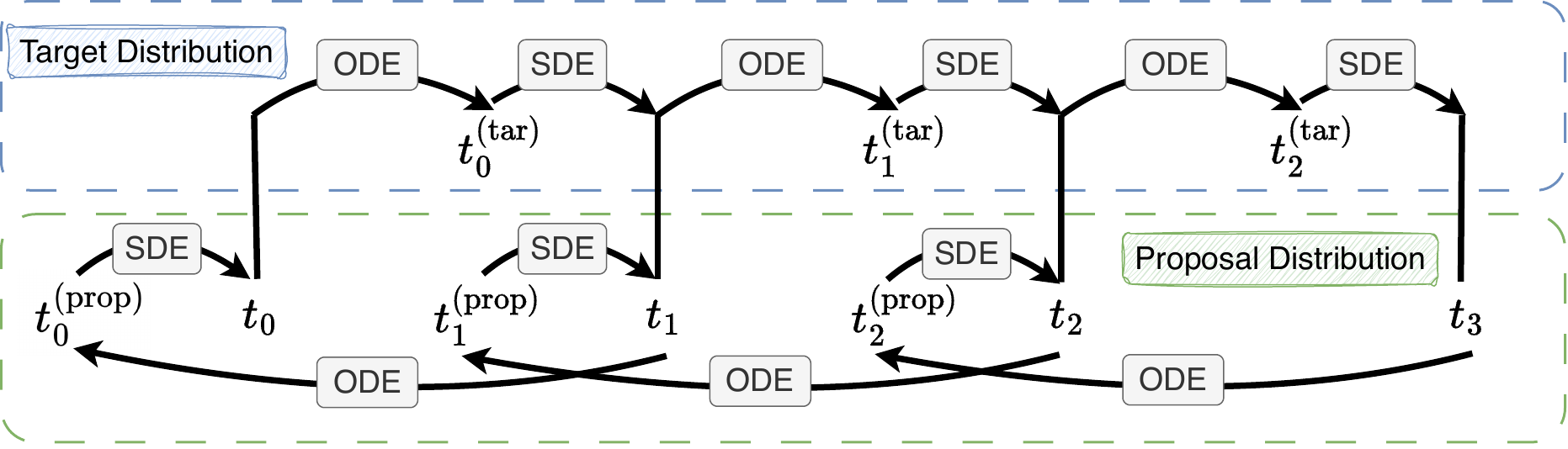}
    \caption{Overview of our proposed sampling method, which combines IS with BCMs. \vspace{-5pt}}
    \label{fig: our method intuition}
\end{figure}
\vspace{-4pt}
\section{Method}
\vspace{-3pt}
To combine Importance Sampling (IS) with Diffusion Models (DMs), a straightforward way is to introduce sequential target and proposal distributions. 
Specifically, for DDPM using $N$ time steps, we can define proposal and target in the joint space for $\boldsymbol{x}_{t_{0:N}}$ at all time steps.
We include a detailed description of this baseline in Appendix \ref{appendix: baseline}.
While producing unbiased samples, this method is computationally inefficient, requiring $N \approx 100$ even for simple targets. Reducing $N$ increases sample errors and proposal-target distribution deviations, negatively impacting IS effectiveness.
\par
To address these challenges, we introduce Consistency Models (CMs) to reduce the number of steps needed. 
However,  CMs only provide a deterministic mapping and hence cannot be directly used for IS.
To yield proposal samples with well-defined densities for IS, we add some Gaussian noise to the CM output. 
We can view CM as a deterministic traversal along the denoising PF ODE and view the additional Gaussian noise as a short traversal along the diffusion SDE.
As for the target, we can simply add Gaussian noise according to the diffusion SDE to each time step. 
\par
However, this approach still faces efficiency issues due to the mismatch between the proposal and the target. 
This arises because the proposal, defined by CMs and SDEs, is a sequence of conditional Gaussians with nonlinear transformations over the means, while the target distribution is simply a sequence of linear conditional Gaussians.
To mitigate this mismatch, a natural solution is to redefine the target using a similar nonlinear transformation, applied inversely. 
Thus, we modify our target distribution: we first move deterministically along the PF ODE in reverse (i.e., toward the diffusion direction), and then add noise according to the diffusion SDE to ensure a valid density. As illustrated in Appendix \ref{appendix: compare}, this design achieves better alignment between the target and the proposal.
\par
Since our method requires efficient traversal along the PF ODE in both forward and backward directions, BCMs naturally emerge as the ideal model for achieving this bidirectional traversal. 
Figure \ref{fig: our method intuition} provides an overview of this process and we detail the proposal and target in the following:

\paragraph{Proposal Distribution.} The proposal distribution is defined as 
$p(\boldsymbol{x}_{t_N})\prod_{n=1}^N p_{\boldsymbol{\theta}}(\boldsymbol{x}_{t_{n-1}}|\boldsymbol{x}_{t_n})$ where
\begin{equation}
\label{equation: proposal n=1}
p_{\boldsymbol{\theta}}(\boldsymbol{x}_{t_{n-1}}|\boldsymbol{x}_{t_n}) = \mathcal{N}\left(\boldsymbol{x}_{t_{n-1}}; f_{\boldsymbol{\theta}}\left(\boldsymbol{x}_{t_n}, t_n\rightarrow t^{(\text{prop})}_{n-1}\right), \left(t^2_{n-1} - \left(t^{(\text{prop})}_{n-1}\right)^2\right)\boldsymbol{I}\right),
\end{equation}
for $n = 1, \dots, N$, where $t^{(\text{prop})}_{n-1}<t_{n-1}<t_n$. Here, $f_\theta$ represents the BCM mapping $\boldsymbol{x}_{t_n}$ from time $t_n$ to $t^{(\text{prop})}_{n-1}$ along the ODE trajectory. Then, noise is added according to the diffusion SDE to move forward to $t_{n-1}$.
For the last time step, unlike conventional definitions, we fix $t_0^{(\text{prop})}$ at $\epsilon$, and tune $t_0$ as a hyperparameter. Samples at $t_0$ are then returned for importance sampling.

\paragraph{Target Distribution.} Similarly, the target distribution also uses the ODE and SDE framework. For a sample at time $t_n$, we first map it forward to $t_n^{(\text{tar})}$ along the PF ODE trajectory. Then, we add noise according to the diffusion SDE to reach $t_{n+1}$. Denoting the true distribution as $\pi$, the target distribution is defined as $\pi(\boldsymbol{x}_{t_0})\prod_{n=1}^Nq_{\boldsymbol{\theta}}(\boldsymbol{x}_{t_{n}}|\boldsymbol{x}_{t_{n-1}})$, where:
\begin{equation}
q_{\boldsymbol{\theta}}(\boldsymbol{x}_{t_n}|\boldsymbol{x}_{t_{n-1}}) = \mathcal{N}\left(\boldsymbol{x}_{t_n}; f_{\boldsymbol{\theta}}\left(\boldsymbol{x}_{t_{n-1}}, t_{n-1}\rightarrow t^{(\text{tar})}_{n-1}\right), \left(t^2_n - \left(t^{(\text{tar})}_{n-1}\right)^2\right)\boldsymbol{I}\right),
\end{equation}
for $n = 1, \dots, N$, where $t_{n-1} \leq t^{(\text{tar})}_{n-1} < t_n$. 
\par
\paragraph{Time Step Optimization.} Having the forms of the proposal and the target, we still need to determine $t_n$, $t_n^{(\text{tar})}$, $t_n^{(\text{prop})}$ for each $n$ (except for $t_0^{(\text{prop})}$ and $t_N$, which are fixed to be $\epsilon$ and $T$ respectively).
In our experiments, we tune these hyperparameters by minimizing the forward KL divergence between target and proposal distributions.  We include details on the hyperparameter tuning in Appendix \ref{appendix: tune time steps}.

\paragraph{Model Extensions.}
To apply our approach to equivariant datasets (e.g., molecules), we incorporate EGNNs \cite{satorras2021n, hoogeboom2022equivariant} to achieve E(3)-equivariant DMs and (B)CMs.
Specifically, following \cite{xu2022geodiff,hoogeboom2022equivariant}, we maintain equivariance by defining target and proposal distributions on the zero-center-of-gravity linear subspace for particle datasets.
However, we observed that standard CM and BCM perform poorly when enforcing equivariance.
In contrast, we found CTM \citep{kim2023consistency} with E(3)-equivariance to be more accurate, likely due to its unique parameterization, which can distill the teacher PF ODE more accurately.
Therefore, we extend CTMs to Bidirectional CTMs (BCTMs).
While our empirical results show that BCM and BCTM do not differ significantly in GMM toy experiments, BCTMs achieve better performance for equivariant potentials. 
Appendix \ref{appendix: bctm} provides further details.

\section{Experiments}
\begin{figure}[t]
    \centering
    \includegraphics[width=.98\linewidth]{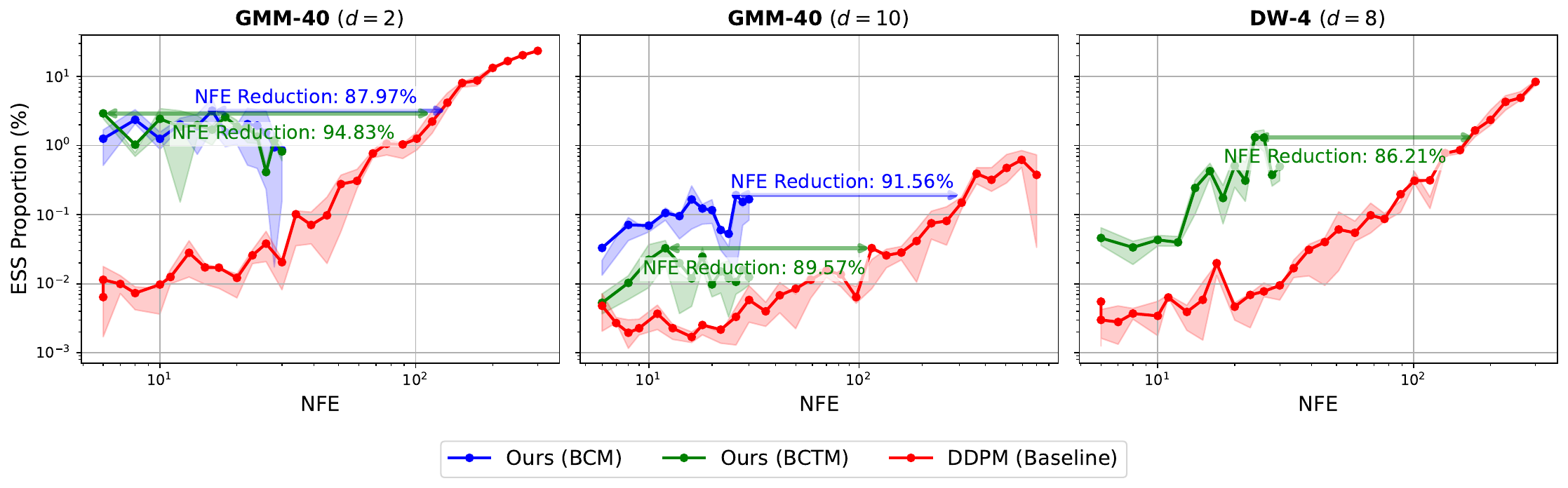}
    \caption{ESS of our proposed algorithm with BCM and BCTM. We also include DDPM as baseline. We estimate ESS with 100,000 samples and report the mean, the first and the third quantile. \vspace{-10pt}}
    \label{fig: ess compare}
\end{figure}

We evaluate our algorithm on synthetic and equivariant $n$-body system datasets, including a 40-component Gaussian Mixture Model (GMM) in 2D and 10D, and a 4-particle double-well potential (DW-4) in 8D. 
For GMM targets, we train score-based DMs and BCM/BCTM using MLPs, following \cite{karras2022elucidating}'s preconditioning and parameterization methods. 
For DW-4, we employ E(3)-equivariant DMs and BCTMs. 
Due to training challenges with equivariant BCMs, we report only BCTM results for this dataset.
Specifically, we perform two evaluations:

\textbf{ESS Comparison.} We report the ESS estimated by samples from the proposal to assess the efficiency of importance sampling. Figure \ref{fig: ess compare} shows the ESS results for our proposed algorithm using both BCM and BCTM, as well as the baseline algorithm.
Notably, compared to the baseline, our method reduces the NFEs required to achieve the same level of ESS by about 85\%.

\textbf{Integral Estimation.} 
Only looking at \emph{estimated} ESS can be misleading. When the target is broader than the proposal, the ESS estimated from finite proposal samples can appear high, even though importance sampling (IS) can present significant errors.

Therefore, we also evaluate the estimation of the integral of some specific test functions $\phi$.
The results are summarized in Table \ref{tab: integral estimation}. More detailed results are included in Appendix \ref{appendix: more experiment results}.
As we can see, our algorithm effectively corrects the model error and achieves a similar performance as the baseline algorithm with much fewer NFEs.
\begin{table}[t]
\centering
\caption{Integral estimates using 100,000 samples, averaged over 5 runs (mean $\pm$ std dev). True value from Monte Carlo sampling of target distribution. Bracketed numbers indicate NFE used.}
\label{tab: integral estimation}
\resizebox{\textwidth}{!}{
\begin{tabular}{lccccccccc}
\toprule
\multirow{2}{*}{Test $\phi(\boldsymbol{x})$} & \multicolumn{3}{c}{GMM-40 ($d = 2$)} & \multicolumn{3}{c}{GMM-40 ($d = 10$)} & \multicolumn{3}{c}{DW-4 ($d = 8$)} \\
\cmidrule(r){2-4} \cmidrule(r){5-7} \cmidrule(r){8-10}
 & True Value & DDPM$+$IS (100) & BCM$+$IS (12) & True Value & DDPM$+$IS (300) & BCM$+$IS (24) & True Value & DDPM$+$IS (150) & BCTM$+$IS (24) \\
\midrule
$\log\|\boldsymbol{x}\|_2$ & 
$3.183\text{\scalebox{0.7}{$\pm 0.00$}}$ & 
$3.174\text{\scalebox{0.7}{$\pm 0.01$}}$ & $3.186\text{\scalebox{0.7}{$\pm 0.01$}}$ & $4.247\text{\scalebox{0.7}{$\pm 0.00$}}$ & $4.247\text{\scalebox{0.7}{$\pm 0.02$}}$ & $4.258\text{\scalebox{0.7}{$\pm 0.02$}}$ & $1.638\text{\scalebox{0.7}{$\pm 0.00$}}$ & $1.641\text{\scalebox{0.7}{$\pm 0.01$}}$ & $1.639\text{\scalebox{0.7}{$\pm 0.00$}}$\\
$\log\|\boldsymbol{x}\|_1$ & 
$3.448\text{\scalebox{0.7}{$\pm 0.00$}}$ & $3.438\text{\scalebox{0.7}{$\pm 0.01$}}$ & $3.451\text{\scalebox{0.7}{$\pm 0.01$}}$ & $5.264\text{\scalebox{0.7}{$\pm 0.00$}}$ & $5.265\text{\scalebox{0.7}{$\pm 0.03$}}$ & $5.275\text{\scalebox{0.7}{$\pm 0.02$}}$ & $2.510\text{\scalebox{0.7}{$\pm 0.00$}}$ & $2.514\text{\scalebox{0.7}{$\pm 0.01$}}$ & $2.511\text{\scalebox{0.7}{$\pm 0.01$}}$\\
$\cos\left(\|\boldsymbol{x}\|_2\right)$ & $0.076\text{\scalebox{0.7}{$\pm 0.00$}}$ & $0.080\text{\scalebox{0.7}{$\pm 0.01$}}$ & $0.078\text{\scalebox{0.7}{$\pm 0.02$}}$ & $0.005\text{\scalebox{0.7}{$\pm 0.00$}}$ & $-0.045\text{\scalebox{0.7}{$\pm 0.03$}}$ & $0.035\text{\scalebox{0.7}{$\pm 0.05$}}$ & $0.382\text{\scalebox{0.7}{$\pm 0.00$}}$ & $0.397\text{\scalebox{0.7}{$\pm 0.03$}}$ & $0.387\text{\scalebox{0.7}{$\pm 0.02$}}$\\
\bottomrule
\end{tabular}
}
\end{table}

\section{Conclusions and Limitations}
In this work, we propose a sampling algorithm that integrates Importance Sampling (IS) with Consistency Models (CMs), enabling unbiased sampling with only a handful of NFEs.
Our method largely outperforms the baseline under limited computational budgets, demonstrating the potential for efficient applications. 
 However, unlike DDPM, which can use more Number of Function Evaluations (NFE) to achieve a higher Effective Sample Size (ESS), our method tends to plateau when the NFE exceeds 10-20.
Moreover, we tuned hyperparameters using the forward KL but found it less effective in higher-dimensional spaces. 
Future work can focus on designing better trade-offs between NFE and performance, as well as identifying more effective hyperparameter tuning metrics.

\section*{Acknowledgements}
JH and JMHL acknowledge support from a Turing AI Fellowship under grant EP/V023756/1.
LIM and JA acknowledge support from Google's TPU Research Cloud (TRC) program. 
LIM acknowledges support from the EPSRC through the Syntech PhD program.

\bibliographystyle{plain}
\bibliography{references} 

\newpage
\appendix

\section{Baseline: Combining Importance Sampling with Diffusion Models}
\label{appendix: baseline}
We now consider task of estimating the integral $\mathbb{E}_{\boldsymbol{x}_0\sim\pi}[\phi(\boldsymbol{x}_0)]$ where $\pi$ is the target distribution and $\phi$ is the test function of interest. We denote the unnormalized target distribution to be $\bar{\pi}$ which we will be able to evaluate. We aim to train a diffusion models to act as the proposal distribution which we can draw samples from. Suppose we have a time interval $[\epsilon, T]$ and we discretize the time interval into $N$ sections with $N+1$ time steps such that $\epsilon=t_0<t_1<\dots<t_N=T$. Assume the conditional proposal distribution of the diffusion model is $p_{\boldsymbol{\theta}}(\boldsymbol{x}_{t_{n-1}}|\boldsymbol{x}_{t_n})$, and the conditional noise distribution is $q(\boldsymbol{x}_{t_n}|\boldsymbol{x}_{t_{n-1}})$ for $n = 1, \dots, N$. Using SNIS, the integral can be estimated as
\begin{align}
\mathbb{E}_{\boldsymbol{x}_0\sim\pi}[\phi(\boldsymbol{x}_0)] &\approx \mathbb{E}_{\boldsymbol{x}_{t_0}\sim\pi}[\phi(\boldsymbol{x}_{t_0})] \\
&= \int \phi(\boldsymbol{x}_{t_0}) \pi(\boldsymbol{x}_{t_0}) d\boldsymbol{x}_{t_0} \\
&= \int \phi(\boldsymbol{x}_{t_0}) \pi(\boldsymbol{x}_{t_0})\left(\prod_{n=1}^N q(\boldsymbol{x}_{t_n}|\boldsymbol{x}_{t_{n-1}})\right)d\boldsymbol{x}_{t_{0:N}} \\
&= \int \phi(\boldsymbol{x}_{t_0}) \left(\frac{\pi(\boldsymbol{x}_{t_0}) \prod_{n=1}^N q(\boldsymbol{x}_{t_n}|\boldsymbol{x}_{t_{n-1}})}{p_{\boldsymbol{\theta}}(\boldsymbol{x}_N) \prod_{n=1}^N p_{\boldsymbol{\theta}}(\boldsymbol{x}_{t_{n-1}}|\boldsymbol{x}_{t_n})}\right)p_{\boldsymbol{\theta}}(\boldsymbol{x}_{t_{0:N}}) d\boldsymbol{x}_{t_{0:N}} \\
\label{equation: importance weights ddpm}
&\approx \frac{1}{K} \sum_{k=1}^K\underbrace{\left( \frac{\pi(\boldsymbol{x}^{(k)}_{t_0}) \prod_{n=1}^N q(\boldsymbol{x}^{(k)}_{t_n}|\boldsymbol{x}^{(k)}_{t_{n-1}})}{p_{\boldsymbol{\theta}}(\boldsymbol{x}^{(k)}_N) \prod_{n=1}^N p_{\boldsymbol{\theta}}(\boldsymbol{x}^{(k)}_{t_{n-1}}|\boldsymbol{x}^{(k)}_{t_n})} \right)}_{w_k}\phi(\boldsymbol{x}_{t_0})
\end{align}
where $\boldsymbol{x}^{(k)}_{t_{0:N}}$ for $k = 1, \dots, K$ are samples from the joint proposal distribution $p_{\boldsymbol{\theta}}$ and $w_k$ for $k=1,\dots, K$ are the importance weights. Note that $\epsilon$ is chosen such that the error for the first approximation is negligible. 

We now need to specify the exact forms of $q(\boldsymbol{x}_{t_n}|\boldsymbol{x}_{t_{n-1}})$ and $p_{\boldsymbol{\theta}}(\boldsymbol{x}_{t_{n-1}}|\boldsymbol{x}_{t_n})$ as well as how to specify the time steps $\{t_n\}$. For the noising distribution, we will define it according to the forward SDE used by \cite{karras2022elucidating}, i.e. $q(\boldsymbol{x}_{t_n}|\boldsymbol{x}_{t_{n-1}})=\mathcal{N}\left(\boldsymbol{x}_{t_n};\boldsymbol{x}_{t_{n-1}}, \left(t^2_n-t_{n-1}^2\right)\boldsymbol{I}\right)$. For the denoising distribution, we can define it in a manner similar to DDIM \citep{song2020denoising}:
\begin{equation}
q_{\boldsymbol{\sigma}}(\boldsymbol{x}_{t_{n-1}}|\boldsymbol{x}_{t_{n-1}}, \boldsymbol{x}_0) = \mathcal{N}\left(\boldsymbol{x}_{t_{n-1}}; \boldsymbol{x}_{t_n}\sqrt{\frac{t^2_{n-1}-\sigma^2_{n-1}}{t^2_n}}+\boldsymbol{x}_0\left(1-\sqrt{\frac{t^2_{n-1} - \sigma^2_{n-1}}{t^2_n}}\right), \sigma^2_{n-1}\boldsymbol{I}\right)
\end{equation}
We define $\sigma_{n-1}$ as $\sigma_{n-1} = \eta\sqrt{\frac{(t^2_n - t^2_{n-1})t^2_{n-1}}{t^2_n}}$ where $\eta \in [0, 1]$. When $\eta=1$, this corresponds to the denoising distribution of DDPM \citep{ho2020denoising}, and when $\eta=0$, the sampling becomes deterministic, corresponding to Euler's first method for solving the PF ODE. Our empirical results indicate that $\eta=1$ consistently yields the highest ESS and lowest IS variance. Therefore, we use the denoising distribution same to that in DDPM \cite{ho2020denoising}.

In our experiments, we adopt the same parameterization, preconditioning, and training methods as introduced by \cite{karras2022elucidating}. Consequently, it is natural to consider the time schedule specified by them, with $t_i = \left(\epsilon^{\frac{1}{\rho}}+\frac{i-1}{N-1}\left(T^{\frac{1}{\rho}} - \epsilon^{\frac{1}{\rho}}\right)\right)^\rho$ and $\rho=7$. However, during subsequent experiments, we observed that increasing the value of $\rho$ results in a higher ESS. When $\rho \rightarrow \infty$, the time schedule approaches: $t_i = \epsilon\left(\frac{T}{\epsilon}\right)^{\frac{i-1}{N-1}}$. This configuration is equivalent to arranging the time steps evenly in logarithmic space. We found that this approach yields the highest ESS across various target distributions when using the baseline method. Therefore, we employ this time schedule in all subsequent experiments related to the baseline method.

\textbf{Strengths:}
The method is grounded in sound theoretical foundations and is straightforward to implement, requiring no modifications to a trained diffusion model. With a sufficient number of steps, it can achieve a high ESS, indicating excellent alignment between the proposal and target distributions. This results in low-variance importance sampling estimates, even in high-dimensional spaces.

\textbf{Limitations:}
The primary weakness of the baseline approach lies in the number of steps required to achieve low-variance estimates. Our experiments reveal that even in one-dimensional space, approximately 100 steps may be necessary to obtain reasonable integral estimates. For higher-dimensional spaces, the required number of time steps increases exponentially, quickly becoming computationally infeasible.

\section{Efficacy of Alternating ODE-SDE Target Distribution Design}
\label{appendix: compare}
\begin{figure}[htbp]
\centering
\begin{subfigure}{\textwidth}
\includegraphics[width=\textwidth]{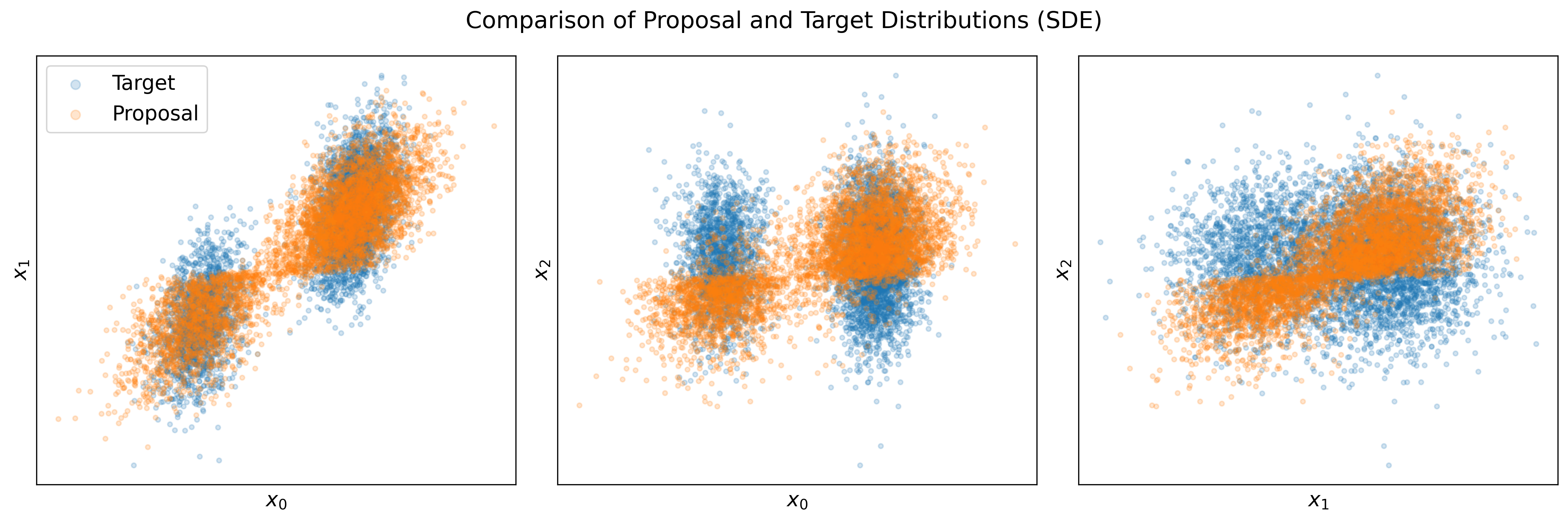}
\end{subfigure}
\begin{subfigure}{\textwidth}
\includegraphics[width=\textwidth]{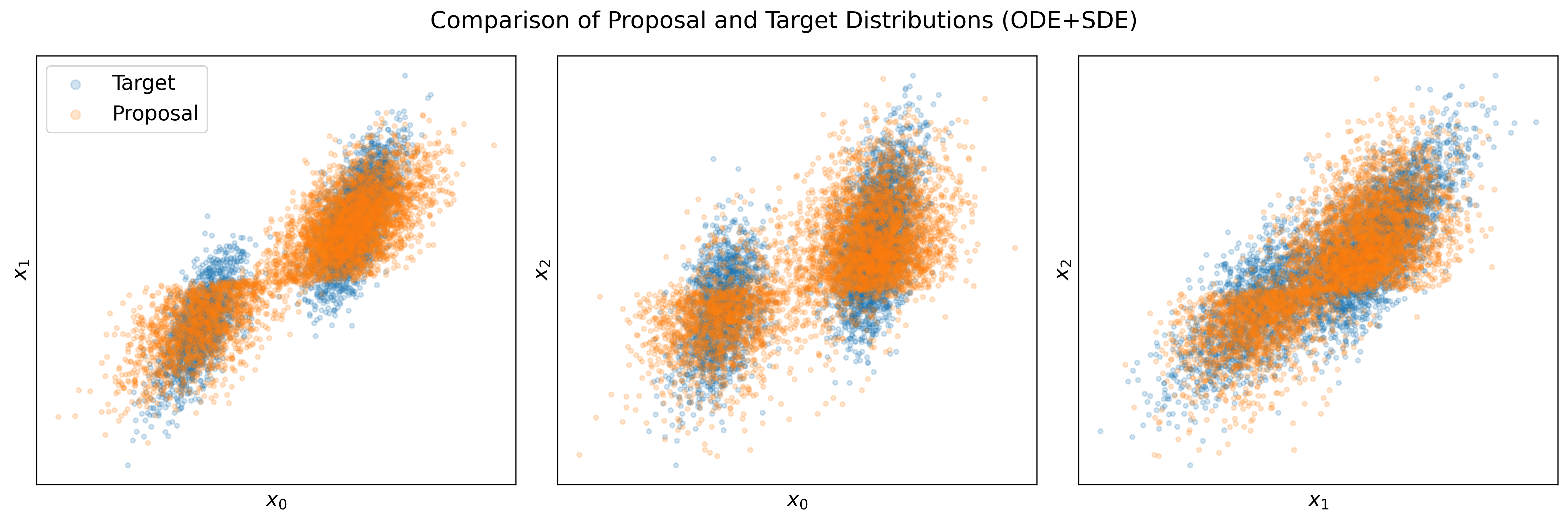}
\end{subfigure}
\caption{Visualization of proposal and target distributions for different target distribution designs.}
\label{fig: compare target}
\end{figure}

To demonstrate the importance of target distribution design in aligning proposal and target distributions for importance sampling, we conducted a comparative study using a simple Gaussian Mixture Model (GMM) with two components as the target distribution. We trained a CTM to learn the target GMM distribution. Setting the number of time steps to three, we optimized the algorithm parameters as detailed in Appendix \ref{appendix: tune time steps} for two scenarios: (1) target distribution modeled by the diffusion SDE alone, and (2) our proposed design with target distribution formed by alternating ODE and SDE. We sampled $\boldsymbol{x}_{t_{0:2}}$ from both proposal and target distributions and visualized them pairwise in Figure \ref{fig: compare target}.

Figure \ref{fig: compare target} demonstrates that combining both ODE and SDE leads to better alignment between proposal and target distributions, validating our hypothesis. It's important to note that this example uses only 3 time steps; increasing the number of time steps will lead to even better alignment, as we will show in the experiment section.

\section{Tuning the Time Steps}
\label{appendix: tune time steps}
Recall that the goal in setting the time steps is to minimize the variance of the IS estimator. Therefore, the key to optimizing our algorithm lies in setting the time steps by minimizing certain objective functions designed to reduce IS variance. To achieve this, we need to introduce a suitable parameterization for the time steps, and we will define three sets of parameters that determine these time steps. For a time interval $[\epsilon, T]$, we reparameterize the three sets of time points as follows:
\begin{align}
\label{equation: time steps}
t_n &= \begin{cases}
T, & n=N\\
\mu_n(t_{n+1} - \epsilon) + \epsilon, & n=0,\dots, N-1
\end{cases}\\
\label{equation: target time steps}
t_{n}^{(\text{tar})} &= (t_{n+1} - t_{n})\eta_{n} + t_{n}, \quad n=0, \dots, N-1\\
\label{equation: proposal time steps}
t_n^{(\text{prop})} &= \begin{cases}
\gamma_n(t_{n} - \epsilon) + \epsilon, & n=1, \dots, N-1\\
\epsilon, & n=0
\end{cases}
\end{align}
Note that in Eq. (\ref{equation: time steps}), the time steps are defined sequentially, with each time step at $n$ falling between the minimum time point $\epsilon$ and the previous time point $t_{n+1}$. A similar approach is used in Eq. (\ref{equation: target time steps}) to define the target time steps, where each target time point is positioned between the time points $t_n$ and $t_{n+1}$. For the proposal time points in Eq. (\ref{equation: proposal time steps}), each proposal time point is positioned between $t_n$ and the minimal time point $\epsilon$, except for the last proposal time point $n=0$, which is fixed to be $\epsilon$.

However, in later experiments, we found that the optimal proposal time points (and $t_0$) are best defined to ensure that the proposal variance equals the target variance whenever possible:
\begin{align}
\label{equation: proposal time points}
t_{n}^{(\text{prop})} &= \begin{cases}
\sqrt{\max\{t_{n}^2+\left(t_{n}^{(\text{tar})}\right)^2-t_{n+1}^2, \epsilon^2\}}, & n = 1, \dots, N-1\\
\epsilon, & n = 0
\end{cases}\\
t_0 &= \min\{\sqrt{\max\{t_{1}^2-\left(t_{0}^{(\text{tar})}\right)^2+\epsilon^2, \epsilon^2\}}, t_{1}^{(\text{tar})}\}
\end{align}
To ensure that the variance matches, we aim to define the proposal time points such that $t^2_n - \left(t^{(\text{tar})}_{n-1}\right)^2 = t^2_{n-1} - \left(t_{n-1}^{(\text{prop})}\right)^2$. Additionally, we enforce the constraint that the proposal time points must be greater than the minimal time point $\epsilon$. The case for $n=0$ differs slightly because of the way that proposal distribution defined for $n=0$ (see Figure \ref{fig: our method intuition}). Here, we apply the constraints that the last time point $t_0$ lies between the minimal time $\epsilon$ and $t_n^{(\text{tar})}$. Notice that $\{\mu_{n}\}$ and $\{\eta_n\}$ are parameters within the range $[0, 1]$. During optimization, we can apply a sigmoid function to enforce this constraint and optimize the parameters to find the optimal values.

\section{Extending Consistency Trajectory Models}
\label{appendix: bctm}
We first review the training method for Consistency Trajectory Models (CTMs) and then describe how it can be extended to be bidirectional. To train CTMs, we focus on scenarios where a pre-trained diffusion model is available. The training loss for CTMs consists of two components: the soft consistency loss and an auxiliary loss to enhance training performance and facilitate the learning of the student model.

\textbf{Soft Consistency Loss:}
Suppose we have an ODE solver, a pre-trained score model $\boldsymbol{\phi}$, a student model $\boldsymbol{G}_{\theta}$, and a teacher model $\boldsymbol{G}_{\theta^-}$ where $\theta^-=\text{stopgrad}(\mu\theta^-+(1-\mu)\theta)$. Given a fixed time interval $[\epsilon, T]$, we first sample a time point $t\in[\epsilon, T]$ and then sample a time point $s\in[\epsilon, t)$. We then sample another time point $u\in[s, t)$. Next, we sample $\boldsymbol{x}_0\sim p_{\text{data}}$ from the data distribution and add noise to $\boldsymbol{x}_0$ to obtain $\boldsymbol{x}_t$ following the forward diffusion SDE.

For the teacher model, we use the ODE solver to move from $t$ to $u$ using the pre-trained score model. We then obtain the sample at time $s$ using the teacher model, which is subsequently mapped to time $t_0=\epsilon$ to obtain $\boldsymbol{x}_{\text{target}}$. The process is summarized as follows:
\begin{align}
\tilde{\boldsymbol{x}}_u &= \text{Solver}(\boldsymbol{x}_t, t, u; \boldsymbol{\phi})\\
\tilde{\boldsymbol{x}}_s &= \boldsymbol{G}_{\boldsymbol{\theta^-}}(\tilde{\boldsymbol{x}}_u, u, s)\\
\boldsymbol{x}_{\text{target}} &= \boldsymbol{G}_{\boldsymbol{\theta^-}}(\tilde{\boldsymbol{x}}_s, s, \epsilon)
\end{align}
where $\epsilon\leq s\leq u < t\leq T$. 

For the student model, we directly map $\boldsymbol{x}_t$ from time $t$ to time $s$ and then map it again to time $\epsilon$. The process is summarized as follows:
\begin{align}
\tilde{\boldsymbol{x}}_s &= \boldsymbol{G}_{\boldsymbol{\theta}}(\boldsymbol{x}_t, t, s)\\
\boldsymbol{x}_{\text{est}} &= \boldsymbol{G}_{\boldsymbol{\theta}}(\tilde{\boldsymbol{x}}_s, s, \epsilon)
\end{align}
The loss function is defined as the distance between $\boldsymbol{x}_{\text{target}}$ and $\boldsymbol{x}_{\text{est}}$:
\begin{equation}
\label{equation: ctm loss}
\mathcal{L}_{\text{CTM}}(\boldsymbol{\theta}; \boldsymbol{\phi}) := \mathbb{E}_{t, s, u, \boldsymbol{x}_0}\mathbb{E}_{\boldsymbol{x}_t|\boldsymbol{x}_0}\left[d(\boldsymbol{x}_{\text{target}}, \boldsymbol{x}_{\text{est}})\right]
\end{equation}

\textbf{Auxiliary Losses:} 
In addition to the soft consistency loss, \cite{kim2023consistency} introduce two auxiliary losses to facilitate student learning: the Denoising Score Matching (DSM) loss and an adversarial loss. The DSM loss is used to enforce that $g_{\boldsymbol{\theta}}(\boldsymbol{x}_t, t, t)$ should act as a denoiser for any $t$. The DSM loss is given by:
\begin{equation}
\mathcal{L}_{\text{DSM}}(\boldsymbol{\theta}) := \mathbb{E}_{\boldsymbol{x}_0, t}\mathbb{E}_{\boldsymbol{x}_t|\boldsymbol{x}_0}\left[\|\boldsymbol{x}_0-\boldsymbol{g}_{\boldsymbol{\theta}}(\boldsymbol{x}_t, t, t)\|^2_2\right]
\end{equation}
In our experiments, we found that the adversarial loss did not significantly improve performance (it was mainly used to enhance image quality in the original paper), so we omit the adversarial loss and focus on the CTM and DSM losses.

\textbf{Extending Consistency Trajectory Models to Bidirectional:} In later experiments, we get inspired by Bidirectional Consistency Models (BCMs), which found that it is not necessary to restrict $s$ to be smaller than $t$. Instead, by relaxing this constraint and only requiring $s \neq t$, the CTM distilled from the pre-trained score model becomes a Bidirectional Consistency Trajectory Model (BCTM). This extension allows us to travel along the PF ODE trajectory not only backward in time but also forward in time using only 1 NFE. We will demonstrate the effectiveness of this extended model in the experiment section.

\section{More Experiment Results}
Table \ref{table: appendix integral estimation results} presents additional results for integral estimation tasks. DDPM + MC directly performs integral estimation using MC with samples from a trained DDPM model. The results show clear bias across all three true distributions compared to true values, validating our motivation for unbiased estimation in practical applications. Both the baseline and our proposed algorithm correct this inherent bias through importance sampling. However, our method achieves similar ESS and integral estimation results with significantly fewer Number of NFE than the baseline.

\label{appendix: more experiment results}
\begin{table}[htbp]
\caption{Integral estimates using 100,000 samples, averaged over 5 runs (mean $\pm$ std dev). True value from Monte Carlo sampling of target distribution. Bracketed numbers indicate the NFE used.}
\centering
\label{table: appendix integral estimation results}
\begin{tabular}{l c c c c}
\toprule
Task $\mathbb{E}_{\pi}[\phi(\boldsymbol{x})]$ & $\log\|\boldsymbol{x}\|_2$ & $\log\|\boldsymbol{x}\|_1$ & $\cos(\|\boldsymbol{x}\|_2)$ & ESS (\%)\\
\midrule
\textbf{GMM-40} ($d=2$) &&&&\\
True Samples $+$ MC &
$3.183\pm 0.002$ &
$3.448\pm 0.002$ &
$0.076\pm 0.002$ &
N/A\\
DDPM (100) $+$ MC &
$3.121\pm 0.002$ &
$3.381\pm 0.002$ &
$0.087\pm 0.003$ &
N/A \\
DDPM (100) $+$ IS &
$3.174\pm 0.008$ &
$3.438\pm 0.008$ &
$0.080\pm 0.013$ &
$2.6\pm 1.1$\\
BCTM (12) $+$ IS &
$3.186\pm 0.009$ &
$3.451\pm 0.010$ &
$0.078\pm 0.015$ &
$2.8\pm 1.3$
\\\midrule
\textbf{GMM-40} ($d=10$)  &&&&\\
True Samples $+$ MC &
$4.247\pm 0.001$ &
$5.264\pm 0.001$ &
$0.005\pm 0.003$ &
N/A\\
DDPM (300) $+$ MC &
$4.149\pm 0.001$ &
$5.147\pm 0.001$ &
$0.000\pm 0.002$ &
N/A \\
DDPM (300) $+$ IS &
$4.247\pm 0.021$ &
$5.265\pm 0.027$ &
$-0.045\pm 0.025$ &
$0.2\pm 0.1$\\
BCTM (24) $+$ IS &
$4.258\pm 0.017$ &
$5.275\pm 0.020$ &
$0.035\pm 0.048$ &
$0.2\pm 0.1$\\\midrule
\textbf{DW-4} ($d=8$) &&&&\\
True Samples $+$ MC &
$1.638\pm 0.000$ &
$2.510\pm 0.000$ &
$0.382\pm 0.000$ &
N/A\\
DDPM (150) $+$ MC &
$1.614\pm 0.000$ &
$2.481\pm 0.000$ &
$0.283\pm 0.001$ &
N/A \\
DDPM (150) $+$ IS &
$1.639\pm 0.004$ &
$2.513\pm 0.004$ &
$0.389\pm 0.017$ &
$1.2\pm 0.2$\\
BCTM (24) $+$ IS &
$1.640\pm 0.005$ &
$2.511\pm 0.006$ &
$0.391\pm 0.020$ &
$1.2\pm 0.5$ \\
\bottomrule
\end{tabular}
\end{table}

\end{document}